\documentclass[sigconf]{acmart}
\AtBeginDocument{%
  \providecommand\BibTeX{{%
    \normalfont B\kern-0.5em{\scshape i\kern-0.25em b}\kern-0.8em\TeX}}}

\usepackage{tabularx} 
\usepackage{booktabs} 
\usepackage{comment}
\usepackage{soul}
\usepackage{caption}
\usepackage{float}
\usepackage{amsfonts}
\usepackage{multirow}
\usepackage{hyperref}
\usepackage{amsfonts}
\usepackage{amsmath,bm}
\usepackage{amsthm}
\usepackage{algorithm}
\usepackage{algcompatible}
\usepackage{enumitem}

\def\BibTeX{{\rm B\kern-.05em{\sc i\kern-.025em b}\kern-.08emT\kern-.1667em\lower.7ex\hbox{E}\kern-.125emX}}

\copyrightyear{2021}
\acmYear{2021}
\setcopyright{iw3c2w3}
\acmConference[WWW '21]{Proceedings of the Web Conference 2021}{April 19--23, 2021}{Ljubljana, Slovenia}
\acmBooktitle{Proceedings of the Web Conference 2021 (WWW '21), April 19--23, 2021, Ljubljana, Slovenia}
\acmPrice{}
\acmDOI{10.1145/3442381.3449987}
\acmISBN{978-1-4503-8312-7/21/04}

\newcommand{\bE}{{\mathbb{E}}}

\newcommand{\Ho}{{\mathcal{H}_{\textrm{on}}}}

\definecolor{mygreen}{rgb}{0.0, 0.37, 0.14}

\begin{document}

\title{GuideBoot: Guided Bootstrap for Deep Contextual Bandits\\in Online Advertising}


\author{Feiyang Pan}
\affiliation{
  \institution{Institute of Computing Technology, Chinese Academy of Sciences.}
  \city{Beijing}\country{China}
}
\authornote{At Key Lab of Intelligent Information Processing of Chinese Academy of Sciences~(CAS). Also at University of Chinese Academy of Sciences~(UCAS).}

\author{Haoming Li}
\affiliation{%
\institution{Institute of Computing Technology, Chinese Academy of Sciences}
\city{Beijing}\country{China}
}\authornotemark[1]

\author{Xiang Ao}
\affiliation{%
\institution{Institute of Computing Technology, Chinese Academy of Sciences}
\city{Beijing}\country{China}
}\authornotemark[1]
\authornote{\,Correspondence to: Xiang Ao <aoxiang@ict.ac.cn>.}

\author{Wei Wang}
\affiliation{%
 \institution{Tencent}
\city{Shenzhen}\country{China}
}

\author{Yanrong Kang}
\affiliation{%
 \institution{Tencent}
\city{Shenzhen}\country{China}
}

\author{Ao Tan}
\affiliation{%
 \institution{Tencent}
 \city{Shenzhen}\country{China}
}

\author{Qing He}
\affiliation{%
\institution{Institute of Computing Technology, Chinese Academy of Sciences}
\city{Beijing}\country{China}
}\authornotemark[1]

\renewcommand{\shortauthors}{Pan, et al.}
\begin{abstract}
The exploration/exploitation (E\&E) dilemma lies at the core of interactive systems such as online advertising, for which contextual bandit algorithms have been proposed. \emph{Bayesian} approaches provide guided exploration via uncertainty estimation, but the applicability is often limited due to over-simplified assumptions. Non-Bayesian \emph{bootstrap} methods, on the other hand, can apply to complex problems by using deep reward models, but lack a clear guidance to the exploration behavior. It still remains largely unsolved to develop a practical method for complex deep contextual bandits. 

In this paper, we introduce Guided Bootstrap (GuideBoot), combining the best of both worlds. GuideBoot provides explicit guidance to the exploration behavior by training multiple models over both real samples and noisy samples with fake labels, where the noise is added according to the predictive uncertainty. The proposed method is \emph{efficient} as it can make decisions on-the-fly by utilizing only one randomly chosen model, but is also \emph{effective} as we show that it can be viewed as a non-Bayesian approximation of Thompson sampling. Moreover, we extend it to an online version that can learn solely from streaming data, which is favored in real applications. Extensive experiments on both synthetic tasks and large-scale advertising environments show that GuideBoot achieves significant improvements against previous state-of-the-art methods.
\end{abstract}



\maketitle

\section{Introduction}
With the progress of deep learning, machine-learning-powered online advertising and recommender systems have achieved significant success in recent years for its ability to model complex deep dependencies between users, ads, and contexts. However, for cost-sensitive decision-making problems such as the cold-start problem, it is still challenging to develop practical methods for minimizing the cumulative regret. 

Online advertising can be viewed as a massive decision-making problem, where the ultimate goal is to deliver the right ads to the right users. The trade-off between exploration and exploitation (E\&E) is crucial to solve this problem, as the decision-maker can receive feedback only through interaction with the environment. Such a problem can be formulated as contextual bandits, for which many algorithms have been proposed and proved successful either theoretically or empirically in specific settings. 

There are two major tracks for exploration in contextual multi-armed bandits: Bayesian and non-Bayesian methods. Bayesian methods provide \textit{guided} exploration by estimating the model uncertainty with Bayesian posterior inference and make decisions thereby. For example, for linear bandits where the dependencies between inputs and rewards are linear functions, LinUCB~\cite{li2010contextual} and Thompson Sampling~\cite{chapelle2011empirical} rely on Bayesian Linear Regression (BLR) to infer the posterior reward distribution. For Bernoulli contextual bandits, GLM-UCB~\cite{filippi2010parametric} uses Bayesian logistic regression to derive the policy. There are also studies to use Gaussian Processes \cite{krause2011contextual} and Neural Processes \cite{garnelo2018neural} for posterior inference in non-linear settings. However, these methods are difficult to scale because exact probabilistic inference is often intractable in complex problems with high-dimensional inputs and non-linear dependencies between features and outcomes. 

In contrast, non-Bayesian methods do not require a specific functional form of the underlying model, thus are suitable for complex problems. A typical choice is the \textit{bootstrap} heuristic, i.e., to train multiple reward models (i.e., deep neural networks) with different initialization \cite{osband2016deep} as well as on randomized subsets of samples \cite{tang2015personalized}, and to randomly use one of them to make a decision. However, the behavior of such a bootstrap exploration method is not explicitly guided. The sub-policy induced by each single model is essentially greedy, so there is a big chance that the overall policy eventually degenerates to a sub-optimal greedy policy. 
Therefore, it still remains largely unsolved to develop practical methods for complex deep contextual bandits. 

In this paper, we propose a novel contextual bandit algorithm named Guided Bootstrap (GuideBoot for short), which combines the best of both Bayesian and non-Bayesian methods. The high-level idea of GuideBoot is to provide explicit uncertainty-based guidance to the bootstrap resampling behavior. 

Similar to traditional bootstrap-based methods, our method trains a bag of reward models and randomly chooses one to make a decision at each step. In contrast to previous methods, we not only use different subsets of collected samples to train the models, but also randomly generate a small number of fake samples while training the models, so as to enable distributional predictions and preclude overfitting. Particularly, every fake sample is generated randomizing the reward signal of a real sample, and the probability of generating such a fake sample is in proportion to the predictive uncertainty of the reward. In this way, unfamiliar contexts and rarely chosen actions can have chance to be explored.

\begin{figure}[t]
\begin{center}
\includegraphics[width=0.99\columnwidth]{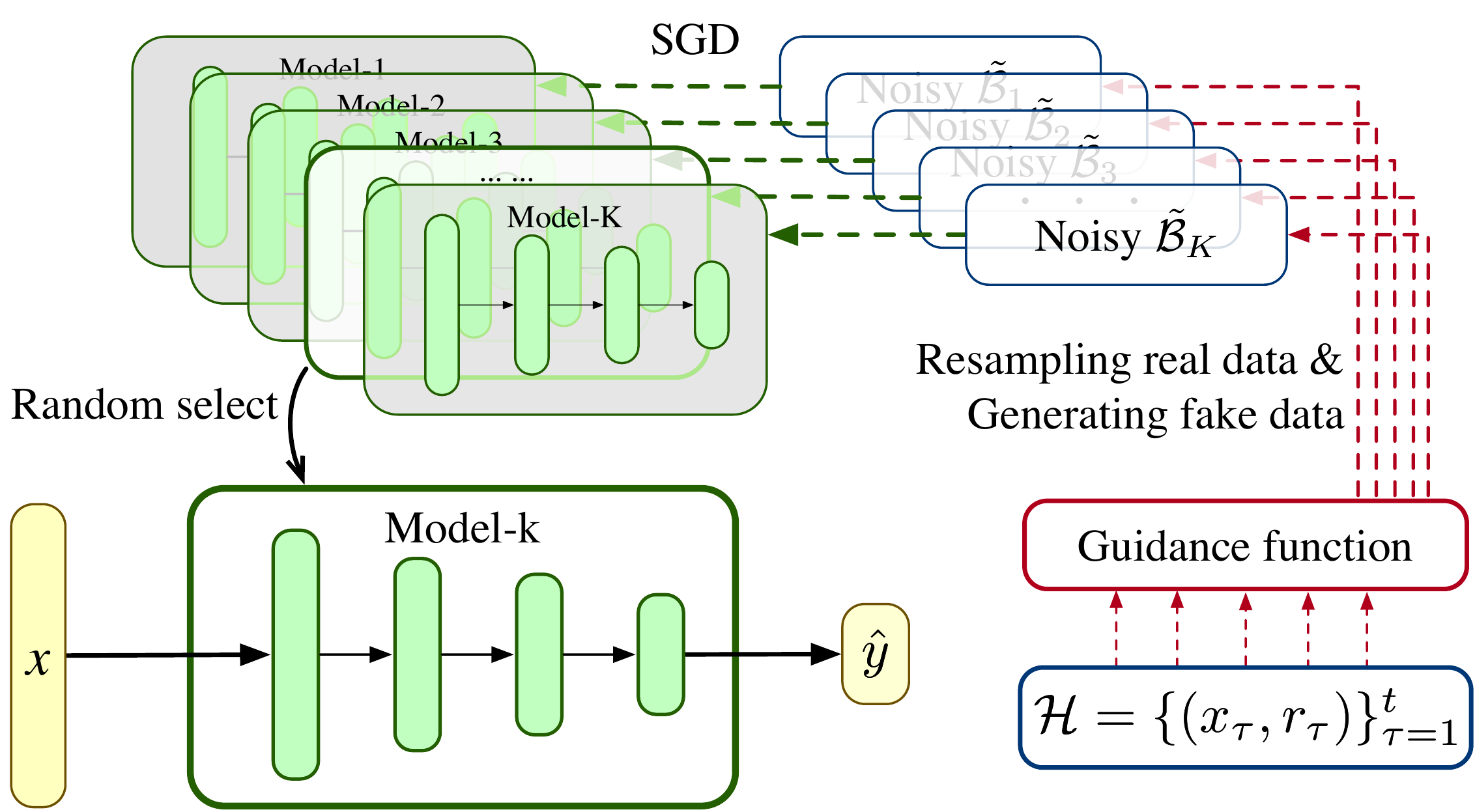}
\caption{Inference and training of GuideBoot. The solid lines are feed-forward passes for inference, making use of one model out of $K$ models. The dashed lines are for training, including a data generation process for each model (red dashed lines) and back-propagation to train each model (green dashed lines).}
\label{fig:GuideBoot}\end{center}
\end{figure}

An essential benefit of GuideBoot is that it can always make decisions on the fly with minimal computational complexity during inference. As shown in Figure \ref{fig:GuideBoot}, it leaves the time-consuming uncertainty estimation to the training stage rather than the inference (decision-making) stage. Therefore, GuideBoot can be incorporated with arbitrary reward models (i.e., modern deep neural networks for click-through rate prediction) and any type of uncertainty estimation approaches ranging from simple non-parametric count-based methods to sophisticated Bayesian methods. 

To better adapt to real applications of high-frequency decision-making such as online advertising and recommender systems, we put forward an extended version of GuideBoot that can handle streaming data efficiently, namely Online GuideBoot. The biggest challenge for developing the online algorithm is that the agent can only learn from streaming data without accessing a complete history. Our algorithm replaces the bootstrap resampling step with a shuffle-and-split step which operates solely on an online buffer, which is easy to implement and empirically powerful.

We conduct on systematical experiments on synthetic tasks and two large-scale real-world tasks on industrial applications of online advertising. The experimental results demonstrate that our methods perform better and are more stable than the state-of-the-arts in both classic settings and online settings.



The remainder of this paper is organized as follows. Section~\ref{sec:preliminary} introduces the problem settings and backgrounds. Section~\ref{section:guideboot} details GuideBoot in a standard setting where the agent maintains a complete history. Section~\ref{section:guideboot-online} illustrates how to deploy our algorithm in an online setting with streaming data, where the agent only learns from recent experiences without accessing to the whole history.
Section~\ref{sec:exp} demonstrates the experiments. 

\section{Preliminaries}\label{sec:preliminary}
\subsection{Contextual multi-armed bandits}
We consider a standard contextual multi-armed bandit problem \cite{langford2008epoch,li2010contextual}. At each step, the environment reveals a \emph{context} (i.e., the user attributes in a recommender system) and $m$ candidate actions (i.e., items in the platform). Without loss of generality, we write $x_{t,a}\in\mathcal{X}\subset\mathbb{R}^d$ as input features of the $a$-th action at step $t$, including the relevant context information. The environment also defines the expected reward for each action by a stationary function $
y_{t,a}\triangleq\bE[r_t(a)] = f^*(x_{t,a})$, where the reward function $f^*(\cdot)$ is unknown to the agent. The agent is required to choose one action $a_t$ (i.e., an item to recommend) out of $m$ candidates $\{1,\dots,m\}$. For simplicity, we write $x_t \triangleq x_{t,a_t}$. Then, the environment draws a reward according to $x_{t}$. As many applications on the web involve optimizing binary outcomes (i.e., clicks or conversions), we assume the reward is drawn from a Bernoulli distribution, i.e., $r_t \sim \textrm{Bernoulli}(y_{t})$, where $y_t \triangleq f^*(x_{t}) \in [0,1]$ is the expected reward. We put a general assumption that $f^*(\cdot)$ is stationary when deriving the GuideBoot algorithm in Section \ref{section:guideboot}. This stationary assumption can be relaxed when extends to the online version in Section \ref{section:guideboot-online}.
\subsection{Bayesian approaches}
Bayesian methods for contextual bandits rely on accurate probability inference of the expected reward so as to obtain an uncertainty estimation for each action. Let $\mathcal{D}_{t-1}=\{(x_\tau, r_\tau)\}_{\tau=1}^{t-1}$ be the collected samples before time step $t$. A Bayesian method not only learns an approximate reward model $\hat{y}_a = f_\theta(x_a)$, but also learns the posterior distribution of the parameters $p(\theta|\mathcal{D}_{t-1})$. By using the posterior, UCB-type algorithms can derive the upper confidence bound $\hat{y}^{\textrm{UCB}}_a$ of the expected reward, i.e.,
$p(y_a < \hat{y}^{\textrm{UCB}}_a)>1-\delta,$
and choose the action with the highest upper confidence bound
$$a_t = \arg\max_a \hat{y}^{\textrm{UCB}}_a.$$
Thompson sampling, on the other hand, randomly samples a set of parameters from the posterior $\theta'\sim p(\theta|\mathcal{D}_{t-1})$ to predict the reward, and make a decision thereby,
$$a_t = \arg\max_a f_{\theta'}(x_a).$$

The drawbacks for these Bayesian methods are two-folds. First, exact probabilistic inference is often intractable for real problems. When the reward function is assumed as a (generalized) linear model, the posterior can be inferred with Bayesian linear (logistic) regression. For complex problems, it is infeasible especially when one would like to use a deep neural net to predict the reward. 

Second, even if we assume the posterior is known, there is a substantial increase in computing burden required for making a decision than a simple greedy policy. For linear bandits, LinUCB \cite{li2010contextual} and GLM-UCB~\cite{filippi2010parametric} need a time complexity of $O(d^2)$ to compute the upper confidence bounds, an order of magnitude slower than an ordinary linear model to make predictions. Thompson sampling with linear payoffs requires $O(d^3)$ to sample the parameter from the multivariable Guassian posterior distribution. In non-linear cases, the Gaussian Process bandit \cite{krause2011contextual} requires even more computations to get the kernel matrix. Recent Bayesian neural nets that rely on Monte Carlo Variational Inference \cite{kingma2015variational,blundell2015weight,gal2016dropout,gal2017concrete} also significantly increase the inference time cost. Such slowdown in decision-making is often unacceptable in real applications. In this paper, the proposed method has the same computational complexity as a standard feed-forward neural net, which can make decisions on-the-fly.

\subsection{Non-Bayesian approaches}
On the contrary, non-Bayesian methods can be applied to problems with arbitrary complex reward functions. For example, the simplest heuristic $\epsilon$-greedy chooses the \texttt{argmax} action $\arg\max_a f_{\theta}(x_a)$ at probability $1-\epsilon$, and explores randomly otherwise. Boltzmann policy computes the \texttt{softmax} over predicted rewards of candidate actions to derive a stochastic policy, which is also shown competitive in \cite{pan2019policy}. Bootstrap-based exploration is also shown effective in both reinforcement learning \cite{osband2016deep,pan2020trust} and bandit problems \cite{eckles2014thompson,tang2015personalized,elmachtoub2017uai,vaswani2018new,pmlr-v97-kveton19a}
, which either maintain multiple bootstrap samples of the history or train multiple reward models from different subsets of data. These methods can be incorporated with deep neural networks, therefore can achieve state-of-the-art performance in deep contextual bandits.
\citet{tang2015personalized} use a Poisson distribution to draw duplicated samples in order to mimic the resampling procedure with online streaming data. \citet{elmachtoub2017uai} proposed a contextual bandit algorithm that follows the same bootstrapping way but uses decision trees as the base learner. 
Apart from bootstrapping real samples and updating with real observed rewards, \citet{pmlr-v97-kveton19a} proposed Giro that designs pseudo rewards for each pull of the arm. \citet{wang2020residual} bounded the pseudo rewards with the perturbed residual from the history. 

However, these methods lack explicit guidance of an explicit notion of uncertainty or prior, therefore they tend to have sub-optimal regret comparing to the Bayesian methods in non-deep settings. Our work is orthogonal to these unguided exploration methods as our GuideBoot perturbs the bootstrapped samples under an explicit guidance of the uncertainty of the reward, where the guidance is achieved by manipulating fake samples to mimic a data prior. 

\subsection{Other related work}
There are also studies on theoretical justification for bootstrap in the (contextual) multi-armed bandits problem. \citet{osband2015bootstrapped} proposed a bandit algorithm named BootstrapThompson and showed the algorithm approximates Thompson sampling in Bernoulli bandits. \citet{vaswani2018new} generalized it to categorical and Gaussian rewards. \citet{hao2019bootstrapping} extended UCB with multiplier bootstrap and derived both problem-dependent and problem-independent regret bounds for the proposed algorithm. 
Although with strong theoretical properties, these methods often limit to oversimplified settings. 
In this paper, we also provide a Bayesian point of view to explain our algorithm (in Section \ref{sec:bayes}). We show that our GuideBoot can approximate the Bayesian posterior with an informative prior, which is more general and flexible than previous work \cite{osband2015bootstrapped} that is based on a degenerate prior.

\section{GuideBoot} \label{section:guideboot}
\subsection{Overview of the method}
GuideBoot is a practical method for deep contextual bandits that can make decisions on the fly. The basic inference and training procedure is shown in Figure \ref{fig:GuideBoot}. 
In this part, we describe the procedure of model updates \emph{over a logged dataset} (aka the \emph{experience buffer}), as in the same setting with previous studies on bootstrap-based contextual bandits \cite{tang2015personalized,pmlr-v97-kveton19a}. Later in Section \ref{section:guideboot-online} we will introduce a more practical implementation of online GuideBoot with online \emph{streaming} data.

A GuideBoot agent maintains $K$ independently trained reward models $f^{(1)}(\cdot), \dots, f^{(K)}(\cdot)$. At the $t$-th step, when a query comes to the system with a number of candidate actions, the agent randomly selects one model $f^{(k)}(\cdot)$ to predict the scores of the actions and chooses the action with maximum score, i.e.,
\begin{equation}
    a_t = \arg\max_a f^{(k)}(x_{t, a}).
\end{equation} 

For training, all models are updated in parallel with gradient descent over different bootstrapped subsets of data consisting of both real samples and randomly generated fake samples. The pseudo-code of GuideBoot is outlined in Algorithm \ref{alg:guideboot}.

The key novelty of GuideBoot is the uncertainty-based guidance for the trade-off between exploration and exploitation, achieved by training the bag of models with perturbed datasets including a small fraction of fake samples. In the rest of this chapter, we introduce \emph{what} the fake samples are in Section \ref{sec:guideboot-what}, \emph{how} to train the models using the fake samples in Section \ref{sec:guideboot-how}, and \emph{why} the fake samples should be used in Section \ref{sec:bayes}.

\begin{table}[tb]
\centering
\caption{Notations}
\begin{tabularx}{\columnwidth}{l X} 
    \toprule
    Notation & Description \\
    \midrule
    $t$ & the time step \\
    $m$ & the number of candidate actions \\
    $K$ & the number of candidate models \\
    $\mathcal{H}_{1:t}$           & The \textit{replay buffer} that stores all history samples \\
    $\Ho$           & The \textit{online buffer} that temporarily collects a batch of samples before updating models \\
    $x_{t,a}$         & Input features of candidate action $a$ (including the context feature) at step $t$ \\
    $\hat{y}_{t,a}$   & The predicted expected reward of action $a$ at step $t$ \\
    $a_t$             & The selected action at step $t$ \\
    $x_t$               & Simplified notation of $x_{t,a_t}$\\
    $\Tilde{\mathcal{B}}_i$ & The $i$-th resampling subset of samples \\
    $f^{(k)}(\cdot)$         & The $k$-th  reward model \\
    $g(\cdot)$              & The \textit{guidance function} $g$ that maps an input $x$ to a probability of generating fake samples \\
    $\hat\rho(\cdot)$        & The unnormalized density estimator \\
    \bottomrule
\end{tabularx}
\label{tab:notations}
\end{table}

\subsection{Guidance from fake samples}\label{sec:guideboot-what}
We start with an example showing when vanilla bootstrap resampling fails. Consider a cold-start advertising problem where we would like to maximize the click-through rate (CTR). Consider two candidate ads \texttt{Ad\_A} and \texttt{Ad\_B}, whose groundtruth CTRs are 0.02 and 0.03, respectively. Suppose that both ads have already been impressed 50 times, \texttt{Ad\_A} has got 1 click, and \texttt{Ad\_B} has got no click. In this case, unfortunately, the bootstrap estimator for the CTR of \texttt{Ad\_B} will always be naught in both mean and variance. Although the estimation perfectly fits the collected data, the induced policy would always recommend \texttt{Ad\_A}, which is obviously sub-optimal.

To fix this issue, intuitively we should introduce a ``prior knowledge'' to the agent, e.g., ``the CTR for an ad is not likely to be zero''. Then the agent can explore unfamiliar actions following the prior when there is not enough data, and exploit greedily with the learned reward models when there is sufficient data for all candidate actions. Conventionally, parametric Bayesian approaches add the prior distribution to model parameters as a regularization in the training process. In contrast, we achieve this by introducing a prior on the data rather than on the parameters, and the derived algorithm can be successfully incorporated with non-parametric bootstrap techniques.

The data prior is achieved by generating \textit{fake samples} during training to regularize the optimization process. 
The generation procedure of fake samples is as follows. At step $t$, if the agent selects the action $a_t$ with input features $x_t \triangleq x_{t,a_t}$ and receives reward $r_t$, it collects a real sample $(x_{t}, r_t)$. Then, we generate fake samples $(x_t, 0)$ and $(x_t, 1)$, which have the same input to the real sample but fake rewards. Next, when training the reward model, each of the two fake samples are getting used at probability $0<g(x_{t})\leq 1$, and are leaved unused otherwise.

By adding the randomized fake samples, the sample distribution can be smoothed, so it is less likely that the agent would be over-confident on its prediction.
We call the probability of generating the fake samples $g(\cdot): \mathcal{X}\rightarrow (0,1]$ a \emph{guidance function} or a \emph{guidance}, which is essentially the salience weight on the data prior given the context $x_t$. 
Intuitively, the guidance function represents the uncertainty about the input: the more the agent is familiar with the action, the smaller the probability is given by the guidance function. This can be achieved by using simple count-based methods or sophisticated Bayesian methods. A formal probabilistic explanation as well as the specific choices of the guidance function for real applications are detailed in Section \ref{sec:bayes}.

Let's come back to the example in the beginning of this section. By using the data prior, if we assign a positive guidance function to generate fake samples for both \texttt{Ad\_A} and \texttt{Ad\_B}, the bootstrapping estimator by resampling from the data for \texttt{Ad\_B} can be non-zero with a large probability, so there is always a chance for \texttt{Ad\_B} being impressed.

\begin{algorithm}[tb]
\caption{GuideBoot (Experience Replay)}
\label{alg:guideboot}
\begin{algorithmic}[1] 
\REQUIRE Number of models $K$, bootstrap batchsize $b$.
\REQUIRE Parameter $\alpha$ for the guidance $g(\cdot)$ in Eq. (\ref{eq:g-cb}).
\STATE Initialize Replay Buffer $\mathcal{H}=\emptyset$;
\STATE Initialize $K$ models $f^{(1)}, \dots, f^{(K)}$ independently;
\STATE Initialize the density function $\hat{\rho}$.
\FOR{$t=1, 2, \dots$}
\STATEx \quad  // Part 1. Decision-making  
\STATE Observe $m$ actions $x_{t,1},\dots ,x_{t,m}$.
\STATE Choose $k\sim \mathcal{U}\{1,K\}$ as the model index.
\STATE Estimate $\hat{y}_{t,a} \gets f^{(k)}(x_{t,a})$ for all $a=1,\dots,m$.
\STATE Take action $a_t \gets \arg\max_{a} \hat{y}_{t,a}$ and receive reward $r_t$.
\STATE Append the history $\mathcal{H} \gets \mathcal{H} \cup \{(x_{t,a_t}, r_t)\}$.
\STATEx \quad // Part 2. Prepare bootstrap data
\STATE Update the density function $\hat{\rho}$ with $(x_{t,a_t}, r_t)$.
\FOR{$k=1,\dots,K$}
\STATE $\mathcal{B}_k \gets$ resample a batch with size $b$ from $\mathcal{H}$.
\STATE $\tilde{\mathcal{B}}_k \gets \mathcal{B}_k$
\FOR{each record $(x_j, r_j)$ in $\mathcal{B}_i$}
    \STATE $g_j \gets \min\{\alpha / \hat\rho(x_j), 1\}$
    \STATE With probability $g_j$:\par
        \hskip\algorithmicindent\quad  $\tilde{\mathcal{B}}_k \gets \tilde{\mathcal{B}}_k \cup \{(x_j,1)$\}\par
    \STATE With probability $g_j$:\par
        \hskip\algorithmicindent\quad  $\tilde{\mathcal{B}}_k \gets \tilde{\mathcal{B}}_k \cup \{(x_j,0)$\}\par
\ENDFOR
\ENDFOR
\STATEx \quad // Part 3. Update the models
\FOR{$k=1,\dots,K$}
    \STATE Update model $f^{(k)}$ with gradient descent on $\Tilde{\mathcal{B}}_k$
\ENDFOR
\ENDFOR
\end{algorithmic}
\end{algorithm}

\subsection{Training with resampled real samples and generated fake samples}\label{sec:guideboot-how}

The training of GuideBoot combines vanilla bootstrap resampling from real data and generated fake samples. 

We need to train $K$ models (e.g., $K$ feed-forward neural networks) over an experience buffer $\mathcal{H}_{1:t}=\{(x_\tau, r_\tau)\}_{\tau=1}^t$ of all the collected samples up to time step $t$. During training, all the models are randomly initialized and then updated in parallel with mini-batch gradient descent. For an update of the $k$-th model, we first follow the Bag of Little Bootstraps (BLB, \cite{kleiner2014scalable}) to perform a $b$-out-of-$t$ bootstrap from $\mathcal{H}_{1:t}$ (i.i.d. resampling with replacements) and get a subset $\mathcal{B}_{k}$, where $b$ is a hyper-parameter for the size of $\mathcal{B}_{k}$. Next, for each sample $x\in \mathcal{B}_{k}$, we compute the guidance function $g(x)$ and generate fake samples $(x, 0)$ or $(x,1)$ at probabilities $g(x)$. So we can get a mixed set of samples $\tilde{\mathcal{B}}_k$ as the union of $\mathcal{B}_{k}$ and the set of generated fake samples. Finally, we perform a step of gradient descent to minimize the negative log-likelihood over the samples $\tilde{\mathcal{B}}_k$ to update the model. 

The pseudo code of GuideBoot is shown in algorithm \ref{alg:guideboot}. As different models are trained with different set of samples, the bootstrap resampling, fake data generation, and gradient descent updates for the bag of models can be done in parallel. 

\subsection{GuideBoot as a Bayesian approximation}\label{sec:bayes}
It is known in the statistics literature that ``the bootstrap distribution represents an (approximate) nonparametric, \emph{noninformative} posterior'' (see chapter 8.4 of \cite{hastie2009elements}), which is also used in \cite{osband2015bootstrapped} to derive a bootstrap-based multi-armed bandit. In this part, we put forward a novel result that our GuideBoot can be viewed as an approximation of an \emph{informative} posterior due to the explicit guidance.

To specifically show the connection between GuideBoot and Bayesian Inference for our bandit task, let's first consider a simplified multi-armed bandit problem, where we have collected a set of feedbacks $r_{a,1}, \dots, r_{a,n}\sim \textrm{Bernoulli}(y_a)$ for an action $a$, with $0\leq n_s\leq n$ successes out of $n$. In this case, if we use Bayesian inference to estimate $y_a$ by $\hat{y}_a$ with a symmetric prior distribution of $\textrm{Beta}(\alpha, \alpha)$, the posterior distribution of $\hat{y}_a$ is 
\begin{equation}
    \hat{y}_a\mid \{r_{a,1}, \dots, r_{a,N}\} \sim \textrm{Beta} (\alpha+n_s, \alpha+n-n_s),
\end{equation}
which is a biased estimator of $y_a$ with a mean of $\frac{\alpha+n_s}{2\alpha+n}$ and a variance of $\frac{(\alpha+n_s)(\alpha+n-n_s)}{(2\alpha+n)^2(2\alpha+n+1)}$. 

On the other hand, the bootstrap mean estimator obtained by resampling $n$ i.i.d. samples with replacement from this data follows a binomial distribution
\begin{equation}
    n\hat{y}_a^{b} \sim \textrm{Bi}(n, n_s/n)
\end{equation}
where $\textrm{Bi}(\cdot,\cdot)$ denotes the binomial distribution. So $\hat{y}_a^{b}$ is an unbiased estimator of $y_a$ with a mean of $n_s/n$ and a variance of $n_s(n-n_s)/n^3$. So the mean and variance of the posteriors of $\hat{y}_a$ and $\hat{y}_a^b$ are very similar. 

Notice that the bootstrap posterior can be viewed as an approximate Bayesian posterior with a noninformative prior $\alpha\rightarrow 0$. However, in bandit algorithms, an informative prior is essential, e.g., classic Thompson Sampling for Bernoulli multi-armed bandits often uses a prior of $\textrm{Beta}(1, 1)$ for each arm. So now the problem is, how do we modify the bootstrap algorithm to approximate a Bayesian posterior with an informative prior?

The solution is GuideBoot. In the above case, we can achieve this by adding two fake feedbacks 0 and 1 to the data and then run bootstrap resampling. Specifically, the bootstrap weights on the fake rewards are set as $\alpha$. The resulted estimator $\hat y^g_a$ (superscript $g$ for GuideBoot) has a mean of $\frac{\alpha+n_s}{2\alpha+n}$ which is the same as $\hat y_a$, and a variance of $\frac{(\alpha+n_s)(\alpha+n-n_s)}{(2\alpha+n)^3}$ which is asymptotically the same as that of $\hat y_a$.

Further, as bootstrap resampling is with replacement, we show that $\hat y^g_a$ can also be constructed by adding fake samples \emph{after} a standard bootstrap resampling procedure. That is, for each resampled data point, we add the fake samples with rewards 0 or 1, both at a probability $\alpha/n$. Since $n$ is the number of impression counts of action $a$, we can rewrite this probability as 
\begin{equation}\label{eq:g-mab}
    g(a) = \alpha / \textrm{Count}(a),
\end{equation}
which is the guidance function of GuideBoot when applying to non-contextual multi-armed bandits. With the analysis above, such an algorithm can be viewed as a nonparametric approximation to the Bayesian Thompson Sampling with Beta prior. Note that moving the fake data generation procedure after resampling is essential in our algorithm, which enables experience replay without adding fake samples to the buffer. It also enables online training as described in Section \ref{section:guideboot-online}.

Finally, the methodology described above can be extended to contextual bandits. When each sample has a featured input $x$ and the reward models are deep neural networks, it is non-trivial to use Bayesian methods because neither the prior nor the posterior of the neural nets can be obtained easily. However, our nonparametric GuideBoot is still simple to run. Recall that the $\textrm{count}(a)$ in Eq. (\ref{eq:g-mab}) can be viewed as an unnormalized density estimation of action $a$. With the same intuition, the guidance function for contextual bandit with input $x$ is formulated as a measure of uncertainty which is inversely proportional to an unnormalized density estimator,
\begin{equation}\label{eq:g-cb}
    g(x) = \alpha / \hat\rho(x).
\end{equation}
where $\hat\rho(x)$ denotes an unnormalized density estimation for input $x$ in the experience buffer. In particular, we assume that $\hat\rho(x)\geq1$ for any seen input $x$ so that the value of $g(x)$ is strictly bounded. 

Now we recommend several specific choices of $\hat{\rho}(\cdot)$ as well as the guidance function. From the analysis above, one can see that the mentioned count-based guidance function in Eq. (\ref{eq:g-mab}) becomes a special case of Eq. (\ref{eq:g-cb}) by letting $\hat\rho(x_a)=\textrm{Count}(a)$, where $a$ can be a categorical index inside the context vector $x_a$. For example, for cold-start advertising where we would like to promote exploration on small ads or new ads, we can simply let 
\begin{equation}
    \hat\rho(x_a) = \textrm{Count}(\texttt{ad\_ID}),\label{eq:rho-count}
\end{equation} where $\textrm{Count}(\texttt{ad\_ID})$ is the count of historical impressions of this ad. For inputs that contain multiple categorical fields, one can alternatively use an unscaled harmonic average of counts, i.e., 
\begin{equation}
    \hat\rho(x) = [\sum_{j=1}^J \textrm{Count}^{-1}(x_j)]^{-1},\label{eq:rho-diag}
\end{equation}
with $J$ the number of categorical fields and $x_j$ the $j$-th entry of $x$. It has shown in previous study on Thompson Sampling \cite{riquelme2018deep} that it works as an Precision Diagonal posterior approximation of Bayesian linear regression which yields good performance in a range of tasks.

\section{Online GuideBoot} \label{section:guideboot-online}
In real applications such as advertising and recommendation, the algorithm has to respond to large number of concurrent requests and update the models with streaming data in real time. The large amount of data and the demand of real-time response makes the algorithm impossible to do bootstrap resampling from the complete history buffer. Even solely storing the experience buffer can be costly in many applications. To remedy the challenges in online scenario, we propose \emph{Online} GuideBoot, an simple and efficient version of GuideBoot that is more practical for online settings.

\subsection{GuideBoot in the online setting}
\label{sec:online-1}

To better describe the online environment, we give a typical outline of the online bandit using greedy policy in Algorithm~\ref{alg:greedy-online}. The setting of the online learning environment with streaming data is described as follows. We suppose that the online agent always collects a batch of samples before updating itself, which is usual in many real systems with concurrent queries. We denote the collected online mini-batch of $c$ samples by $\Ho\triangleq\mathcal{H}_{t-c:t}$, which should be used to update the reward models. To maximize generality, here the batch-size $c$ can be a varied or fixed number, which depends on the actual design of the online serving. For example, if it is a constant~(e.g., $c=512$), it means that the agent updates its reward model only every $c$ time steps. Likely, $c$ can be a variable, e.g., be the number of online samples received within the last 10 seconds. 

To use bootstrap techniques in such an online setting when there is no complete history data, one can solely perform resampling on the recent batch $\Ho$, which might degrade the performance. However, fortunately, our GuideBoot algorithm can still work well by constructing a randomized dataflow with online streaming data. Although the dataflow we propose does not completely follow the conventional bootstrap resampling, it is strong in empirical performance and easy to implement.

Recall that the reason of using multiple models trained on different resampling datasets is to keep track of the model uncertainty in the predicted rewards. Considering that the reward models~(i.e., logistic regression or neural networks) are usually updated by stochastic gradient descent~(SGD) during real online learning, we can directly get stochastic updates on the models by randomizing the sequence of arriving samples. Therefore, we propose to use shuffling and minibatching instead of resampling to construct the batches of real data. After having the subsets of real data, we use the same process of generating fake samples as in the previous section, which is done independently for each model. Therefore, different models can still receive different fake samples while training.

The pseudo code of the derived online algorithm, named Online GuideBoot, is shown as Algorithm~\ref{alg:guideboot-online}. In particular, after collecting an online buffer $\Ho$ with $c$ samples, we construct the dataflow for updating each model as follows: First, the samples in the buffer is shuffled and split into $n$ disjoint mini-batches, where $n$ is a predefined number of mini-batches. Second, for each mini-batch, we generate fake samples with the same principles described in the previous section and add them into the mini-batches.
Finally, the models are updated by gradient descent on these mini-batches one after another. Therefore, the gradients to update each model is computed over different mini-batches with randomly generated fake samples, which enables stochastic optimization.

\begin{algorithm}[t]
\caption{Conventional greedy algorithm (online)}
\label{alg:greedy-online}
\begin{algorithmic}[1] 
\STATE Initialize the model $f$.
\WHILE{True}
\STATE Clear online buffer $\Ho \gets \emptyset$
\STATEx \quad  // Part 1. Serving
\FOR{$t = 1, \dots, c$}
\STATE Observe $m$ actions $x_{t,1},\dots ,x_{t,m}$
\STATE Estimate $\hat{y}_{t,a} \gets f(x_{t,a})$ for all $a=1,\dots,m$.
\STATE Take action $a_t \gets \arg\max_{a} \hat{y}_{t,a}$ and receive reward $r_t$.
\STATE Append the buffer $\Ho \gets \Ho \cup \{(x_{t, a_t},r_t)\}$.
\ENDFOR
\STATEx \quad  // Part 2. Learning
\STATE Split $\Ho$ into mini-batches $\{\mathcal{B}_i\}_1^n$
\STATE Update model $f$ with gradient descent on $\{\mathcal{B}_i\}_1^n$
\ENDWHILE
\end{algorithmic}
\end{algorithm}

\begin{algorithm}[t]
\caption{Online GuideBoot}
\label{alg:guideboot-online}
\begin{algorithmic}[1] 
\STATE Initialize $K$ models $f^{(1)}, \dots, f^{(K)}$ independently;
\STATE Initialize the density function $\hat{\rho}$.
\WHILE{True}
    \STATE Clear Online Buffer $\Ho \gets \emptyset$
    \STATEx \quad  // Part 1. Serving
    \FOR{$t=1,\dots,c$}
        \STATE Observe $m$ actions $x_{t,1},\dots ,x_{t,m}$.
        \color{mygreen}
        \STATE Choose $k\sim \mathcal{U}\{1,K\}$ as the model index.
        \STATE Estimate $\hat{y}_{t,a} \gets f^{(k)}(x_{t,a})$ for all $a=1,\dots,m$.
        \color{black}
        \STATE Take action $a_t \gets \arg\max_{a} \hat{y}_{t,a}$ and receive reward $r_t$.
        \STATE Append the buffer $\Ho \gets \Ho \cup \{(x_{t, a_t},r_t)\}$.
    \ENDFOR
    \STATEx \quad  // Part 2. Learning
    \color{mygreen}
    \STATE Update the density function $\hat{\rho}$ with $\Ho$.
    \FOR{$k=1,\dots,K$}
        \STATE Shuffle and split $\Ho$ into mini-batches $\{\mathcal{B}_l\}_1^n$.
        \FOR{each mini-batch $\mathcal{B}_l$}
            \FOR{each record $(x_j, r_j)$ in $\mathcal{B}_l$}
                \STATE $g_j \gets \min\{\alpha / \hat\rho(x_j), 1\}$
                \STATE With probability $g_j$:\par
                    \hskip\algorithmicindent\quad\quad $\mathcal{B}_l \gets \mathcal{B}_l \cup \{(x_j,1)\}$\par
                \STATE With probability $g_j$:\par
                    \hskip\algorithmicindent\quad\quad $\mathcal{B}_l \gets \mathcal{B}_l \cup \{(x_j,0)\}$\par
            \ENDFOR
        \ENDFOR
        \STATE Update model $f^{(k)}$ with mini-batches $\{\mathcal{B}_l\}_1^n$
    \ENDFOR
    \color{black}
\ENDWHILE
\end{algorithmic}
\end{algorithm}

\subsection{Implementation and deployment}
\begin{figure}[t]
\centering
\includegraphics[width=\columnwidth]{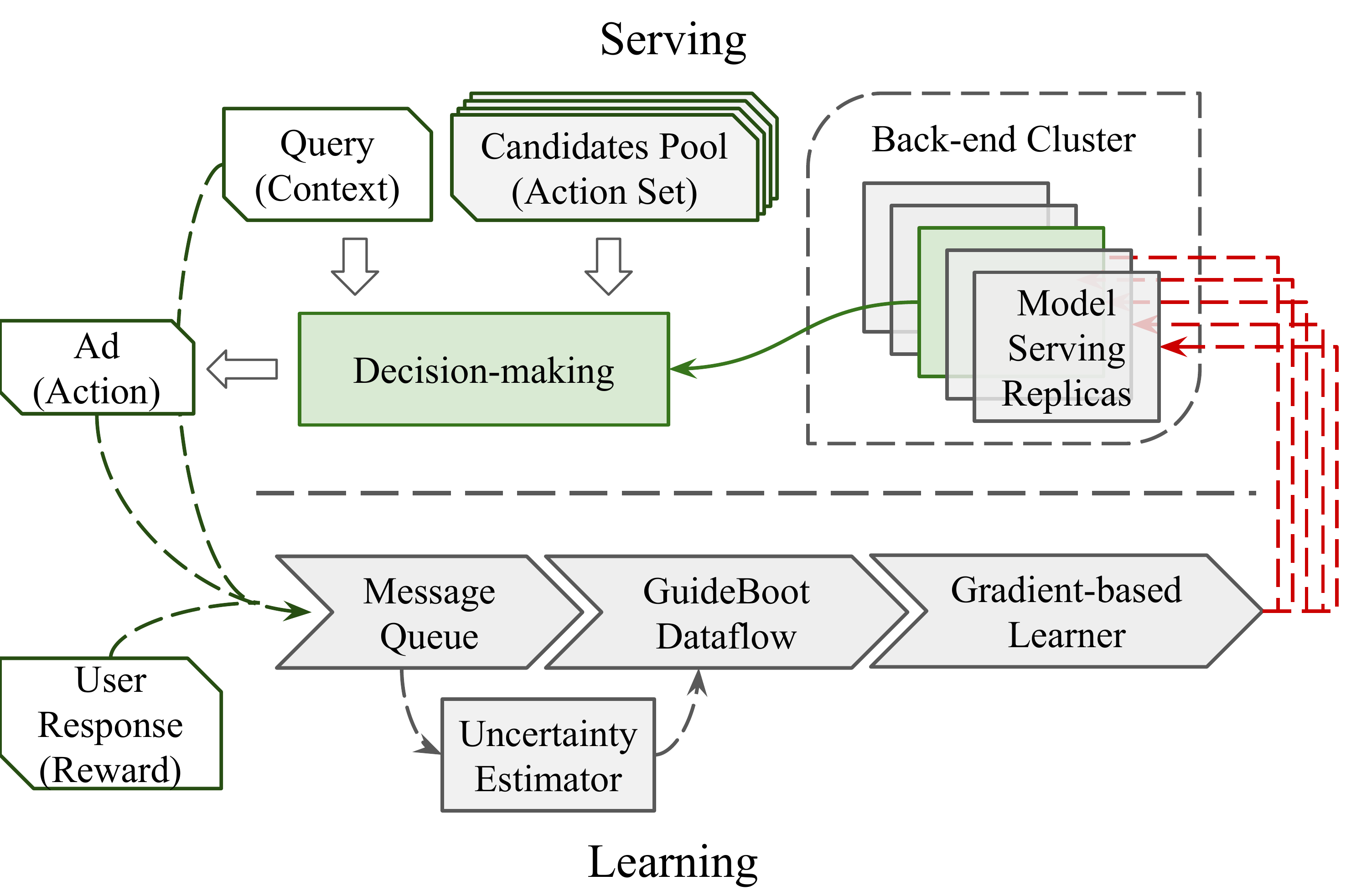}
\caption{Overall pipeline of an online advertising system with Online GuideBoot.}
\label{fig:Deployment}
\end{figure}

Our algorithm can be implemented easily in industrial platforms such as advertising systems \cite{he2014practical,cheng2016wide,mcmahan2013ad} or modern industrial RL platforms \cite{gauci2018horizon}, which is particularly helpful in the cold-start problems \cite{shah2017practical,pan2019warm}. An overview of the learning pipeline on an online advertising system is illustrated in Figure~\ref{fig:Deployment}. The pipeline consists of two parts: online serving and online learning. 

The serving section is responsible for the real-time decision-making for each coming request. The following steps are subsequently processed whenever the system receives a new request with an input context\footnote{We use an online advertising scenario as an example. The input context usually contains a query from a user, along with the text input of the query, the user's attributes, and the user's viewing history.}. 
\begin{enumerate}
    \item[1.] \textbf{Retrieval of the candidate set}. A set of candidate actions (i.e., ads or items) are provided according to the context (query), along with the input attributes and features.
    \item[2.] \textbf{Inference}. Each candidate is evaluated by the reward model randomly selected from the bag of $K$ models and gets a predicted score. The scores are usually the expected rewards, such as the eCPM (expect cost per thousand impressions), which need to be estimated according to the predicted click through rates (pCTR). The system will use a specific model to make the predictions on a backend server, which is chosen randomly from the backend cluster of models. 
    \item[3.] \textbf{Decision-making}. Send the top ranked ad to the user.
\end{enumerate}

On the other hand, the learning section collects user feedback and updates the models. We consider logging a small online buffer with $c$ steps so that we can run the GuideBoot algorithm.
\begin{enumerate}
    \item[1.] \textbf{Logging}. The user's response~(e.g., click or not), along with the input context and selected action, is collected and published to the message queue. Records are fetched from the message queue and stored to an online buffer, until there are enough data to train the models.
    \item[2.] \textbf{Updating the uncertainty estimator}. As described in Section~\ref{sec:bayes}, the guidance function $g(\cdot)$ is inversely proportional to an unnormalized density model, so it can be understood as an uncertainty estimator. After observing the online buffer, the estimator should be updated. For example, when using the count-based guidance in Eq.~(\ref{eq:g-mab}), the system needs to maintain a counting table. 
    \item[3.] \textbf{GuideBoot dataflow}. The dataflow is detailed in Section~\ref{sec:online-1} and line 12-20 in Algorithm \ref{alg:guideboot-online}. It constructs $K$ series of mini-batches in parallel by shuffling and splitting the online buffer, and generating fake samples with the guidance function. 
    \item[4.] \textbf{Gradient-based training}. The learner updates the models with the samples from the GuideBoot dataflow. The models are pushed to the serving section only after the updates are finished. After training, the buffer and the dataflow are cleared and ready for receiving new samples.

\end{enumerate}

\section{Experiments}\label{sec:exp}
In this section, we evaluate our GuideBoot on both synthetic task and large-scale real-world advertising environments. Specifically, we seek to answer the following questions through experiments:
\begin{enumerate}
    \item[\textbf{Q1.}] In a standard non-deep Bernoulli setting, how does GuideBoot compare with classic UCB and Thompson Sampling algorithms?
    \item[\textbf{Q2.}] In complex deep contextual bandits, how does GuideBoot compare with recent advanced deep methods?
    \item[\textbf{Q3.}] What contributes to the improvement of GuideBoot? 
\end{enumerate}

\subsection{Experiments on synthetic task}
\subsubsection{Settings} Although the original motivation of this paper is to build practical algorithms for deep complex contextual bandits, we would like to first check the correctness and effectiveness of GuideBoot in the classic setting. Therefore, we setup standard Bernoulli bandit environments where the reward functions are generalized linear models. In such a setting, it has been extensively studied that a number of UCB and Thompson Sampling based algorithms are provably efficient with sub-linear cumulative regret bounds~\cite{filippi2010parametric}. 

Now we describe the detailed setup of the Bernoulli bandits. At each step, we observe $m=25$ actions, each with three categorical attributes $x_a = [a, x_{a,1}, x_{a,2}]$, where $a=1,\dots,25$, $x_{a,1}\sim \mathcal{U}\{1, 5\}$, and $x_{a,2}\sim \mathcal{U}\{1, 5\}$. For each input $x_a$, the reward is drawn from Bernoulli$(y_a)$, where the oracle expected reward is $y_a = \sigma(\bm{w}_0[a] + \bm{w}_1[x_{a,1}] + \bm{w}_2[x_{a,2}] - 1)$ with $\sigma$ the sigmoid function, $\bm{w}_0\in\mathbb{R}^{25}$, $\bm{w}_1\in\mathbb{R}^{5}$, and $\bm{w}_2\in\mathbb{R}^{5}$ the coefficient vectors for the three fields. Here we use the notation $\bm{w}[i]$ denoting the $i$-th entry of an vector $\bm{w}$. We generate 100 randomized environments of this setting, where the coefficients in $\bm{w}_0$, $\bm{w}_1$ and $\bm{w}_2$ in each environment are i.i.d. drawn from $\mathcal{U}(-0.25, 0.25)$.

\subsubsection{Compared methods} \label{sec:classic-methods} As the problem is standard Bernoulli contextual bandits, we compare our GuideBoot with the following classic algorithms. 
\begin{description}

\item[$\epsilon$-greedy:] The simplest unguided algorithm, which selects the action with the greatest expected reward with probability $1-\epsilon$ and explores randomly otherwise. We set $\epsilon=0.1$ which is a common choices in previous studies.

\item[$\epsilon$-greedy (decay):] A variant of $\epsilon$-greedy with decaying $\epsilon$ as $t$ increases. Although as simple as the ordinary $\epsilon$-greedy, it is commonly used as an empirically strong baseline method. We let $\epsilon$ linearly decay from $0.1$ to zero throughout training.

\item[GLM-UCB:] The UCB algorithm with generalized linear model, proposed in \cite{filippi2010parametric}. It relies on estimating the upper confidence bound of the reward prediction with a Bayesian generalized linear model. The coefficient on the exploration bonus is set as $\alpha_t = \sqrt{\log(t+1)}$ as suggested in its original paper.

\item[TS-BLR:] Thompson Sampling with Bayesian generalized linear model \cite{chapelle2011empirical,agrawal2013thompson}, which is commonly used and is shown to achieve state-of-the-art performance in a recent study  \cite{riquelme2018deep}. The reward model is learned with the same generalized linear model in GLM-UCB.

\item[Bootstrap:] The standard bootstrap resampling method with experience replay, which maintains $K$ reward models trained with resampled data from the history buffer.

\item[OBB:] Online Bootstrap Bandit algorithm proposed in \cite{tang2015personalized}, which enables online bootstrap with duplicated samples where the number of duplicates is drawn from a Poisson distribution.

\item[Giro:] A recent bootstrap-based method by \citet{pmlr-v97-kveton19a}, which trains $K$ models on data resampled from a perturbed history buffer. When each new sample is added to the buffer, a pair of pseudo samples are also added with probability $\alpha=0.5$. 
\end{description}

We test both the experience replay version and the online version of GuideBoot (denoted by \textbf{GuideBoot} and \textbf{Online Guideboot}, respectively), with $\alpha=1$ and $\hat\rho(\cdot)$ in Eq. (\ref{eq:rho-diag}). For all the bootstrap-based methods, we set $K$ to 5, which is a sufficient number for bootstrap models as demonstrated in previous work \cite{tang2015personalized,pmlr-v97-kveton19a}.
Among the bootstrap-based methods, Vanilla Bootstrap, Giro, and GuideBoot require storing the whole history, while OBB and Online GuideBoot are online methods without the whole experience buffer. 
\begin{figure}[t]
\centering
\includegraphics[width=\columnwidth]{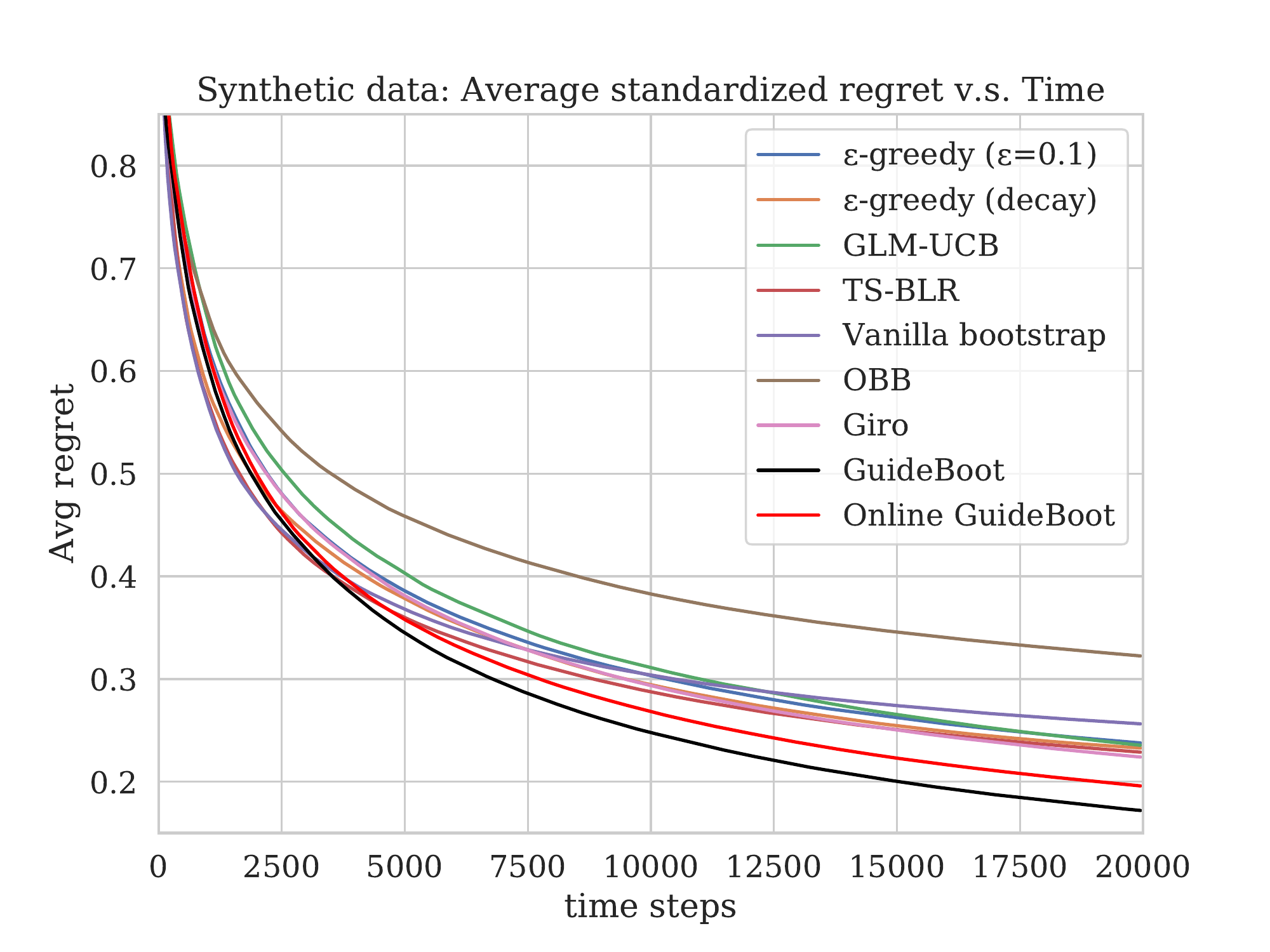}
\caption{Experimental results on the synthetic tasks with 50 random runs. The average regret over time is reported. }
\label{fig:exp1}
\end{figure}
\begin{table}[t]
    \centering
    \caption{Statistics of the Tencent Ads simulator}
    \begin{tabular}{l|c|c}
        \toprule
        Dataset & Tencent Ads A & Tencent Ads B  \\
        \midrule
        \# time steps & 4.5M & 14.7M \\
        \# ads (total) & 8.9K & 6.5K \\
        \# candidates (at each step) & 250 \textasciitilde~450 & 100 \textasciitilde~200 \\
        \# fields of context & 13 & 13 \\
        \# fields of candidate & 12 & 12 \\
        \# features of context & 4.2K & 4.3K \\
        \# features of ad & 12K & 9.4K \\
        \bottomrule
    \end{tabular}
    \label{tab:tencent-simulators}
\end{table}

\subsubsection{Results}
The averaged regret is shown as Figure~\ref{fig:exp1}. From the results, we observe that our GuideBoot works pretty good. Specifically, the experience replay version of GuideBoot performs the best and significantly outperforms the classic Bayesian methods as well as previous bootstrap-based algorithms.

To analyse the results of bootstrap-based algorithms in more detail, we notice that the vanilla bootstrap runs poorly even though it can resample from the whole history. Giro adds a fixed proportion of fake samples to the history and outperforms vanilla bootstrap. 

Next, we find that GuideBoot brings a significant improvement in performance, indicating that the fake sample generation along with the guidance function is essential for the algorithm. By comparing GuideBoot with vanilla bootstrap and Giro, it is clear that the uncertainty-based guidance helps, which answers the question Q3.
\begin{figure*}[t]
\centering
\includegraphics[width=\textwidth]{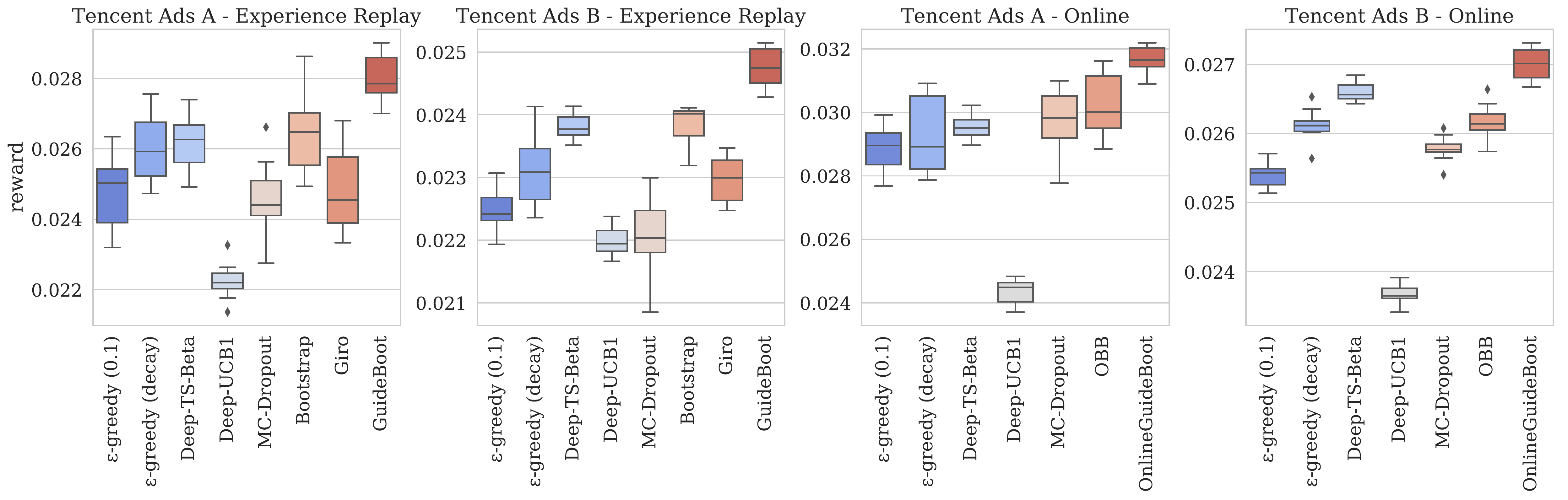}
\caption{The boxplot of the experimental results on advertising tasks with 10 random runs. 
}
\label{fig:exp2}
\end{figure*}

On the other hand, for the bootstrap-based methods learned with streaming data only, OBB as the online extension of the vanilla bootstrap suffers from a huge loss. However, our Online GuideBoot still works fairly well and its relative loss from GuideBoot is small, indicating the effectiveness of the shuffle-and-split step. Online GuideBoot can be a competitive baseline method when it is difficult to access to the whole history.

\subsection{Experiments on online advertising}

Next we conduct experiments on real-world online advertising environments, which involves high-dimensional featured inputs and non-stationary user responses with complex dependencies on the features. 
A large-scale simulator based on historical logs of online advertising platforms is developed for fair comparison due to the infeasibility to deploy online A/B test.


\subsubsection{Advertising simulation setup}

We collected five days of advertising logs from Tencent's online advertising systems. Specifically, we subsampled two logging datasets~(written as \textbf{Tencent Ads A} and \textbf{Tencent Ads B}) from two major sites of Tencent display ads and feed ads. The statistics of the datasets are outlined in Table~\ref{tab:tencent-simulators}.

In the logs, each record is a real request consisting of two parts: the context of the request and a candidate pool. The context includes query and user features such as the user's gender, age, and device type. The candidate pool is a batch of ads with their attributes (e.g., ad ID and advertiser ID). The simulator has the click-through rate (CTR) of each (context, ad) pair given by a sophisticated black-box user model, which is unknown to the agent. When the agent chooses an ad, the simulator samples a binary feedback using the user model. The goal of the agent is to achieve a high average reward throughout learning. 

For each simulator, we construct two sets of experiments: one using experience replay where the model is updated with mini-batches sampled from the whole history, and another using online streaming data without storing all the history. We run all the experiments over 10 random seeds and report the average reward.

\subsubsection{Model structure and compared methods}

The reward function is modeled by a neural network with an embedding layer and two hidden layers. The embedding layer transforms each raw field into a 8-dimensional dense vector. The embeddings of all the input fields are concatenated into a single vector, which is then fed into a ReLU network with two 128-dimensional fully connected layers. The neural net is updated by the Adam optimizer \cite{kingma2014adam} to minimize the Log-Loss (aka negative log-likelihood).

The tested methods are listed as follows. In addition to \textbf{$\epsilon$-greedy}, \textbf{Bootstrap}, \textbf{OBB} and \textbf{Giro} algorithms described in Section \ref{sec:classic-methods}, we also compare the following Bayesian methods that can be incorporated with deep models:

\begin{description}
    \item[Deep-UCB1:] A common method in the industry that combines deep models with UCB1 \cite{auer2002finite}. The uncertainty is estimated from the impression count $\alpha \sqrt{2\log(t)/\textrm{Count}(\texttt{ad\_ID})}$. We set an empirical choice of $\alpha=0.1$.
    \item[Deep-TS-Beta:] Another practical method in the industry that combines Thompson sampling with count-based Beta distributions. The prediction for each input is estimated with Beta$(a, b)$ s.t. $a+b=\textrm{Count}(\texttt{ad\_ID})$ and $a/(a+b)=f(x_a)$. Following \cite{chapelle2011empirical}, we use posterior shaping of $\alpha=0.25$.
    \item[MC-Dropout:] To use Monte-Carlo Dropout during both training and inference. which can be understood as variational Bayes \cite{kingma2015variational,gal2016dropout}. Using dropout for decision-making can be seen as an approximation of Thompson Sampling \cite{gal2016dropout,pan2019policy}. The Dropout is applied only on the last hidden layer \cite{riquelme2018deep} and the Dropout rate is set to 0.1.
\end{description}

For bootstrap-based methods, we again used 5 rewards models. For Giro, we found that large $\alpha$ led to unacceptable performance in the advertising tasks, so we searched over $[0.2,0.1,0.05,0.01]$ and chose $\alpha=0.01$ in this experiment. For our GuideBoot and Online GuideBoot, in order to ensure fair comparison with the count-based Deep-UCB1 and Deep-TS-Beta, we use the count-based density estimator in Eq. (\ref{eq:g-mab}) to construct the guidance function. Note that vanilla Bootstrap and Giro uses the whole history, so they are only tested in the experience replay group. OBB uses online streaming data, so it is tested in the online group.

\begin{table}[tb]
\centering
\caption{Results on Tencent Ads (Experience Replay)}
\begin{tabular}{lccccccccccc}
\toprule
 & \multicolumn{2}{c}{\textbf{Tencent Ads A}} & & \multicolumn{2}{c}{\textbf{Tencent Ads B}} \\
\cline{2-3} \cline{5-6} 
Method & reward & std & & reward & std \\
\midrule
$\epsilon$-greedy ($\epsilon=0.1$)
 & 0.0248    & (0.0010)    & & 0.0225    & (0.0003)    \\
$\epsilon$-greedy (decay)
 & 0.0260    & (0.0009)    & & 0.0231    & (0.0006)    \\
Deep-TS-Beta 
 & 0.0262    & (0.0007)    & & 0.0238    & (0.0002)    \\
Deep-UCB1
 & 0.0222    & (0.0005)    & & 0.0220    & (0.0002)    \\
MC Dropout
 & 0.0245    & (0.0010)    & & 0.0221    & (0.0006)    \\
Bootstrap
 & 0.0264    & (0.0010)    & & 0.0239    & (0.0003)    \\
Giro
 & 0.0249    & (0.0012)    & & 0.0230    & (0.0003)    \\
GuideBoot
 & \textbf{0.0280}    & (0.0006)    & & \textbf{0.0248}    & (0.0003)    \\
 \bottomrule
\end{tabular}
\label{tab:tencent-results-replay}
\end{table}

\begin{table}[tb]
\centering
\caption{Results on Tencent Ads (Online)}
\begin{tabular}{lccccccccccc}
\toprule
 & \multicolumn{2}{c}{\textbf{Tencent Ads A}} & & \multicolumn{2}{c}{\textbf{Tencent Ads B}} \\
\cline{2-3} \cline{5-6} 
Method & reward & std & & reward & std \\
\midrule
$\epsilon$-greedy ($\epsilon=0.1$)
 & 0.0289    & (0.0006)    & & 0.0254    & (0.0002)    \\
$\epsilon$-greedy (decay)
 & 0.0293    & (0.0012)    & & 0.0261    & (0.0002)    \\
Deep-TS-Beta
 & 0.0295    & (0.0004)    & & 0.0266    & (0.0001)    \\
Deep-UCB1
 & 0.0243    & (0.0004)    & & 0.0237    & (0.0001)    \\
MC-Dropout
 & 0.0297    & (0.0010)    & & 0.0258    & (0.0002)    \\
OBB
 & 0.0302    & (0.0009)    & & 0.0262    & (0.0002)    \\
Online GuideBoot
 & \textbf{0.0317}    & (0.0004)    & & \textbf{0.0270}    & (0.0002)    \\
 \bottomrule
\end{tabular}
\label{tab:tencent-results-online}
\end{table}

\subsubsection{Results}
The comparisons are illustrated in Figure \ref{fig:exp2}, and the detailed results are listed in  Table~\ref{tab:tencent-results-replay} and Table~\ref{tab:tencent-results-online}, respectively. 

From the results, we first observe that our GuideBoot and Online GuideBoot have significant and consistent superiority in average cumulative reward comparing to all other baselines. Meanwhile, the standard deviations of GuideBoots are relatively small, indicating that the improvement of our method is robust in all the experiments. 

To provide more insights from the experiments, we find some simple methods are considerably effective, while some sophisticated methods do not always work well. For example, despite its simplicity, the $\epsilon$-greedy algorithm with decaying $\epsilon$ is shown to be strong and can often outperform MC-Dropout. For the Bayesian methods, Deep-UCB1 is much worse than Deep-TS-Beta, most likely because it requires a gradually increasing coefficient for exploration (i.e., the $\sqrt{\log(t)}$ factor) which may cause over-exploration. 

Additionally, in these two simulations, the online algorithms have higher average click-through rate comparing to the experience replay algorithms, which is in contrast with the previous experiments on synthetic data. It is most likely because that the synthetic task has a stationary environment, while in advertising the distributions of inputs and outputs shift over time. Therefore, the online algorithms that updates the models only from recent samples are capable of fast adaptation to the distributional shift. 

Overall, from the experiments on both the synthetic task and complex deep contextual bandits, we find that our GuideBoot and Online GuideBoot methods enjoy both efficiency and effectiveness.

\section{Conclusion and Discussion}\label{sec:conclusion}

In this paper, we propose GuideBoot, an efficient and effective algorithm for addressing the trade-off between exploration and exploitation in deep contextual bandits. GuideBoot trains multiple reward models with resampled subsets of data and guides the exploration behavior by adding fake samples to the resampled subsets. We show that when the probability of generating fake samples is in proportion to the predictive uncertainty of the reward, GuideBoot can be understood as a non-Bayesian approximation to Thompson Sampling with an informative prior. For online applications where resampling from the whole history is infeasible, we propose an online version of GuideBoot that solely learns from the online streaming data, which is efficient and easy to implement in modern platforms. Experiments over synthetic tasks and real-world large-scale advertising show that our method consistently and significantly outperforms previous state-of-the-art methods. 

\begin{acks}
This work is supported by 2019 Tencent Marketing Solution Rhino-Bird Focused Research Program. The research work is also supported by the National Key Research and Development Program of China under Grant No. 2017YFB1002104, the National Natural Science Foundation of China under Grant No. 92046003, 61976204, U1811461, the Project of Youth Innovation Promotion Association CAS and Beijing Nova Program Z201100006820062. 
\end{acks}

\bibliographystyle{ACM-Reference-Format}
\bibliography{bandits}
\end{document}